%% file: root.tex
\documentclass[letterpaper, 10 pt, conference]{ieeeconf}  % Comment this line out if you need a4paper
\usepackage{mathrsfs}
\makeatletter
\let\NAT@parse\undefined

\IEEEoverridecommandlockouts                              % This command is only needed if 
\input{includepackages}
\title{\LARGE \bf
Realizing Learned Quadruped Locomotion Behaviors through Kinematic Motion Primitives 
}

\author{Abhik Singla, Shounak Bhattacharya, Dhaivat Dholakiya, \\  Shalabh Bhatnagar, Ashitava Ghosal, Bharadwaj Amrutur and Shishir Kolathaya% <-this % stops a space
\thanks{This work is supported by the Robert Bosch Center for Cyber Physical Systems, Bangalore, India}% <-this % stops a space
\thanks{Abhik Singla, Shounak Bhattacharya and Dhaivat Dholakiya are with the Robert Bosch Centre for Cyber-Physical Systems, IISc, Bangalore, India. 
        {E-mail: \tt\small \{abhiksingla, shounakb, dhaivatd\}@iisc.ac.in}.}%
\thanks{Shalabh Bhatnagar is with the Faculty of Computer Science and Automation, Ashitava Ghosal is with the Faculty of Mechanical Engineering, Bharadwaj Amrutur is with the Faculty of Electrical \& Computer Engineering, and Shishir Kolathaya is an INSPIRE Faculty Fellow at the Robert Bosch Center for Cyber Physical Systems, IISc, Bengaluru, India
{\tt\small \{shalabh,asitava,amrutur,shishirk\}@iisc.ac.in}}%
}

\begin{document}

\maketitle
\thispagestyle{empty}
\pagestyle{empty}

%%%%%%%%%%%%%%%%%%%%%%%%%%%%%%%%%%%%%%%%%%%%%%%%%%%%%%%%%%%%%%%%%%%%%%%%%%%%%%%%
\begin{abstract}
Humans and animals are believed to use a very minimal set of trajectories to perform a wide variety of tasks including walking. Our main objective in this paper is two fold 1) Obtain an effective tool to realize these basic motion patterns for quadrupedal walking, called the kinematic motion primitives (kMPs), via trajectories learned from deep reinforcement learning (D-RL) and 2) Realize a set of behaviors, namely trot, walk, gallop and bound from these kinematic motion primitives in our custom four legged robot, called the ``Stoch". D-RL is a data driven approach, which has been shown to be very effective for realizing all kinds of robust locomotion behaviors, both in simulation and in experiment. On the other hand, kMPs are known to capture the underlying structure of walking and yield a set of derived behaviors. % In addition, irrespective of the type of gaits obtained from training, the kMPs extracted followed a very similar pattern i.e., the shape of VNMW.
We first generate walking gaits from D-RL, which uses policy gradient based approaches. 
We then analyze the resulting walking by using principal component analysis. %, which is the best available tool for studying walking data. 
% We first train a neural network based policy in simulation, which takes $2-3$ hours of training time. 
% Data obtained from the resulting walking is used to extract the kMPs, which takes only a few seconds. 
We observe that the kMPs extracted from PCA followed a similar pattern irrespective of the type of gaits generated. Leveraging on this underlying structure, we then realize walking in Stoch by a straightforward reconstruction of joint trajectories from kMPs. This type of methodology improves the transferability of these gaits to real hardware, lowers the computational overhead on-board, and also avoids multiple training iterations by generating a set of derived behaviors from a single learned gait.
% Finally we demonstrate walking in Stoch by using a linear combination of these kMPs as reference trajectories. 
%Therefore, in this paper, we show a methodology to use D-RL to generate the kMPs, which can then be used to demonstrate walk, trot, gallop and bound experimentally.
%While the focus is not essentially in obtaining a novel methodology for realizing quadrupedal walking, but to obtain a methodology for principal component analysis  through the well known machine learning techniques.
%but to make a connection with the most recent success behind the reinforcement learning techniques for legged locomotion.
%After the recent success in achieving legged locomotion in both quadrupeds and bipeds, we decided to investigate further

\end{abstract}

%%%%%%%%%%%%%%%%%%%%%%%%%%%%%%%%%%%%%%%%%%%%%%%%%%%%%%%%%%%%%%%%%%%%%%%%%%%%%%%%
\input{Introduction}

%%%%%%%%%%%%%%%%%%%%%%%%%%%%%%%%%%%%%%%%%%%%%%%%%%%%%%%%%%%%%%%%%%%%%%%%%%%%%%%%
% \input{Hardware}

%%%%%%%%%%%%%%%%%%%%%%%%%%%%%%%%%%%%%%%%%%%%%%%%%%%%%%%%%%%%%%%%%%%%%%%%%%%%%%%%

\input{RL}

\input{PCA}

%%%%%%%%%%%%%%%%%%%%%%%%%%%%%%%%%%%%%%%%%%%%%%%%%%%%%%%%%%%%%%%%%%%%%%%%%%%%%%%%
\input{Results}

%%%%%%%%%%%%%%%%%%%%%%%%%%%%%%%%%%%%%%%%%%%%%%%%%%%%%%%%%%%%%%%%%%%%%%%%%%%%%%%%
\section{CONCLUSION} \label{sec: conclusion}
We demonstrated walk, trot, gallop and bound gaits in the quadruped robot Stoch experimentally. The key contribution was in the methodology followed in transferring of the D-RL based gaits from simulation to real hardware. Our approaches were mainly motivated by \cite{moro2013horse} and \cite{moro2012kinematic}, which pointed to the fact that motion primitives have a very similar underlying structure that are independent of the type of gait and the model. %To conclude, we made a honest assessment of the existing approaches existing in literature, and followed an approach that best suited our requirements.
Small variations in gaits and model do not require retraining via D-RL methods from scratch, and kMPs are sufficient for such scenarios. On the other hand, large variations in models and environments require a thorough search, thereby requiring learning based approaches. Since we mainly focused on data driven kMPs in this paper, future work will involve obtaining these motion primitives from continuous functions like polynomials, canonical walking functions or other basis functions that are equally suitable for generating walking  gaits.

%%%%%%%%%%%%%%%%%%%%%%%%%%%%%%%%%%%%%%%%%%%%%%%%%%%%%%%%%%%%%%%%%%%%%%%%%%%%%%%%

%%%%%%%%%%%%%%%%%%%%%%%%%%%%%%%%%%%%%%%%%%%%%%%%%%%%%%%%%%%%%%%%%%%%%%%%%%%%%%%%



\section*{ACKNOWLEDGMENT}
We acknowledge Ashish Joglekar, Balachandra Hegde, Ajay G and Abhimanyu for the PCB design and software development.

\bibliographystyle{IEEEtran}
\bibliography{references}

\end{document}

%% file: includepackages.tex
\usepackage{amsmath,amsfonts,amssymb}
\usepackage{url}
\usepackage{graphicx}
\usepackage{float}
\usepackage{multirow}
\usepackage{tikz}
\usepackage{pgf}
\usepackage{placeins}
\usepackage{subcaption}
\usepackage{comment}
\usepackage{placeins}
\usepackage{latexsym}
\usepackage{pifont}
\usepackage{pgfplots,filecontents}
\usepackage{gensymb}
\usepackage{epstopdf}
\usepackage{siunitx}
\pgfplotsset{compat=newest}

\usepackage[symbol]{footmisc}

\usetikzlibrary{intersections}
\newcommand{\R}{\mathbb{R}}

\newcommand{\E}{\mathbb{E}}

\DeclareGraphicsExtensions{.png,.jpg,.jpeg,.eps,.gif}

%% file: Introduction.tex
\section{Introduction}
Legged locomotion has been well studied for more than six decades, starting from the inception of the GE walking truck in $1965$ \cite{getruck}. A variety of approaches, starting from inverted pendulum models \cite{raibert1986legged}, zero moment points \cite{vukobratovic2004zero}, passivity based control \cite{macgeer,collins2001three}, capture points \cite{prattcapture}, hybrid zero dynamics \cite{hzd_grizzle} to even machine learning based approaches \cite{tedrake2004stochastic,hobbelen_rl_2006} have been explored.
%A more recent approach that has been very powerful is the use of deep neural nets (DNNs) for obtaining robust walking control laws via reinforcement learning.
A more recent trend has been the use of deep reinforcement learning (D-RL) methods \cite{google_paper,DBLP:journals/corr/abs-1803-05580} to determine optimal policies for efficient and robust walking behaviors in both bipeds and quadrupeds. %One such successful approach is by using reinforcement learning to determine the optimal distribution of the actions, given the state of the robot \cite{google_paper,DBLP:journals/corr/abs-1803-05580}.
% To the best of our knowledge, all of the 
D-RL was successfully used to achieve walking in the quadrupedal robot Minitaur \cite{google_paper}, where %successfully demonstrated trot and gallop experimentally. 
the control policy was trained via one of the well known policy gradient based approaches called the proximal policy optimization (PPO) \cite{PPO}. 

% RL for walking is well studied in literature \cite{hobbelen_rl_2006,tedrake2004stochastic} that found a lot of success in simple bipedal walkers (less than $5$ links). Reinforcement learning with DNNs is a more recent trend \cite{google_paper,DBLP:journals/corr/abs-1803-05580} 
% which has also been used in \cite{google_paper}, and more recently in \cite{DBLP:journals/corr/abs-1803-05580}.
% We will be using the same framework to obtain optimal policies for our quadrupedal robot Stoch here, but without prior experimental data which both \cite{google_paper} and \cite{DBLP:journals/corr/abs-1803-05580} used for transferability on real hardware. Our approaches on improving transferability is via kinematic motion primitives (kMPs), which will be discussed in Section \ref{sec:PCA}.

%  \cite{raibert1986legged} showed how to use inverted pendulum based approaches to keep the torso/body upright, thereby preventing {\it falling}. \cite{macgeer} used the passive dynamics of the robot to realize periodic orbits (limit cycles) with the end result being stable walking. \cite{hzd_grizzle} used the notion of {\it hybrid zero dynamics} to describe under-actuations, which, in turn, enabled the much needed transition from traditional flat-footed walking to walking with pointed feet. 

While all the approaches described above were applicable to many kinds of legged robots, there was also an independent group of researchers focusing entirely on capturing symmetries in quadrupedal walking, and then utilize these symmetries to obtain a low dimensional representation for walking \cite{moro2013horse}. In essence, the focus here was more on trajectory reconstruction for each of the joints of the robot from a minimal number of basis functions, called the {\it kinematic motion primitives} (kMPs) \cite{sanger2000human}. It was shown in \cite{moro2012kinematic} that only four-five primitives are enough to represent the various locomotion patterns like walking, trotting, and canter. Comparisons were even made with kMPs derived from horse \cite{moro2013horse}, showing a similarity of the  kMPs irrespective of the gait used. This type of minimal representation of walking point to the fact that the joint angle trajectories do not contain ``high frequency" modulations. 
Alternative representations of these motion primitives by using canonical walking functions \cite{ameshuman}, and $6^{th}$ order Bezier polynomials \cite{hzd_grizzle} also indicate this low dimensional behavior of walking. %In fact, \cite{ameshuman} showed that mean human walking data from the joints also had a high correlation ($>0.99$) with time responses of linear spring-mass-damper systems. %In other words, joint angle trajectories have a high correlation with the time response of linear spring-mass-damper systems.
These observations strongly indicate that a small number of basic trajectories (or basis functions) are sufficient to realize periodic trajectories in the individual joints of the robot. %, which results in walking.

\begin{figure}[t!]
\includegraphics[width = \linewidth]{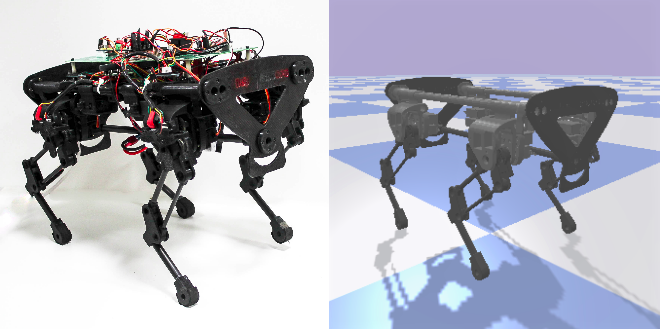}
\caption{Figure showing our custom built quadrupedal robot Stoch, and its simulation model shown in Pybullet simulator.}
\label{fig:robot_pic}
\end{figure}

\begin{figure*}[t!]
\centering
\includegraphics[width=\linewidth]{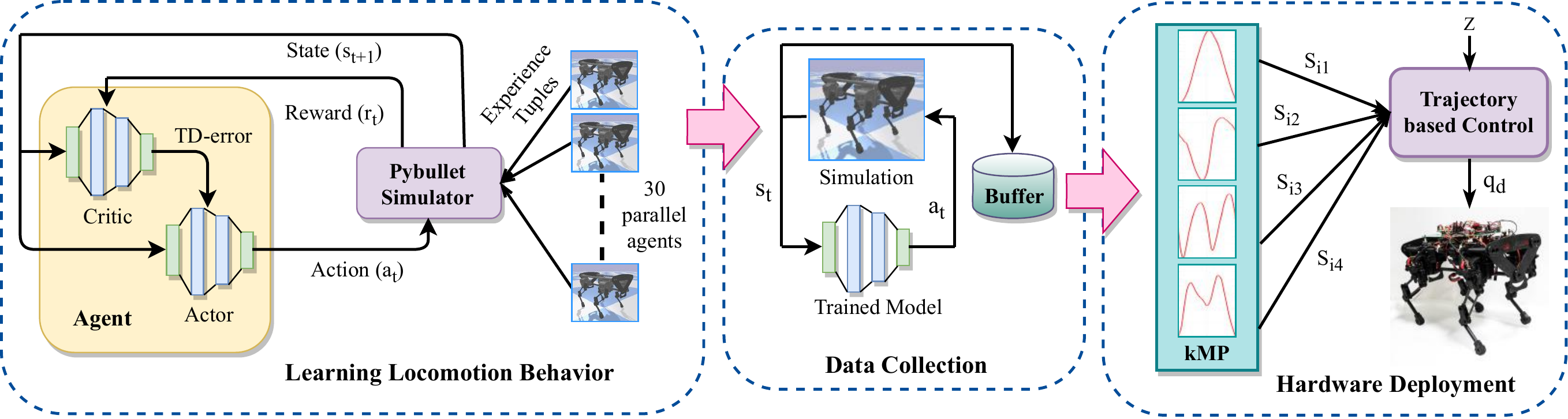}
  \caption{Figure showing a graphical representation of our methodology for realizing quadrupedal walking in Stoch. We use the D-RL based controller for obtaining stable walking gaits in simulation, and then translate them into the quadruped Stoch, experimentally via kinematic motion primitives (kMPs).}
  \label{fig:overall_pic}
\end{figure*}

The walking community is currently divided on deciding which approaches are best for achieving legged locomotion. 
% One side of the community is very formal, i.e., 
Some use formal methods, which are model based and determine the underlying principles of walking, and then develop controllers that yield stable locomotion \cite{macgeer,collins2001three,hzd_grizzle}. %References \cite{macgeer,collins2001three,hzd_grizzle} focused on realizing {\it limit cycle} type of walking in bipeds, which are model based. %It was observed that some of the reference gaits realized were inherently robust to external disturbances \cite{koushil,nyamber}. These reference trajectories draw comparisons with motion primitives and also central pattern generators that are common in biology \cite{ameshuman,ijspeert2008central}. 
Others use model-free methods like RL (and also D-RL) to derive robust locomotion behaviors in both bipeds and quadrupeds \cite{tedrake2004stochastic,hobbelen_rl_2006,google_paper,kohl2004policy}.
% while avoiding the complexity of the dynamics involved in the process. 
% This was the result of the success in realizing dynamic gaits in quadrupeds and bipeds with stochastic gradient descent based approaches. 
% In other words, the goal is more to develop a framework that does not require the understanding of the physics of walking, and simply allow the computer (neural network) to learn the best policy to steer the robot forward all by itself. 
This methodology not only eliminates the difficult and often unpleasant task of identification and modeling of the robot, but also enables the computer to continuously adapt to changing environments. However, it is important to note that the controller learned is model-free, but the training is normally done in simulation which is often not realistic. %In other words, any gap between the actual and the simulation model, called the reality gap, makes the controller ineffective. 
This reality gap is also a well known problem faced by traditional model based optimization tools for walking \cite{rehericradurus,Kuindersma2016,Hereid2017FROST}.

One of the main goals of this paper is to 
% answer some of these questions, and primarily 
approach this problem of reality gap by using the tools used in formal methods. % and, possibly, bridge this divide between the two  communities. 
% Therefore, we would like to study the gaits obtained from D-RL. 
Formal methods usually employ the optimization tools only for generating the reference gaits (trajectories), and then these trajectories are tracked at the joint level by model free methods (e.g. PD control) \cite{rehericradurus,koushil,nyamber}.
Therefore, we would like to view D-RL based training as a trajectory generator (optimization tool) more than the on-board compute platform for walking.
This is similar to the method followed by \cite{kohl2004policy}, where RL was used to find the optimal set of parameters of the foot trajectory. 
% With this setup, we lose the robustness due to feedback, but a quadruped is inherently 
% will first use D-RL to generate gaits in relatively new and unknown models and environments. %, i.e., allow the computer to execute a random search with the highest reward (low power consumption, low energy, minimum time etc.). 
% 
% We will obtain gaits from simulation, and then focus on extracting the kMPs via principal component analysis (PCA). We will compare the trot, pace and bound gaits, and show that all of the kMPs follow a very similar pattern. %With this observation, we establish a methodology to reconstruct walking from these kMPs. 
We will obtain a gait from D-RL, extract kMPs from this gait, and then realize a set of derived behaviors, namely trot, walk, gallop and bound in Stoch experimentally. Trot and gallop were achieved in  \cite{google_paper}  by separate instances of training, while we were able to realize more than four behaviors by one instance of training. A pictorial representation of our approach is given in Fig. \ref{fig:overall_pic}. This hybrid approach allows for a mechanistic way of generating gaits in complex legged locomotion---that combines the data driven D-RL based gait synthesis in simulation leading to extraction of  kMPs which are then used for gait tracking in hardware with minimum computational overhead.

% known technique used in machine learning (cite some paper). Intuitively, PCA is a method to compress a lot of {\it data}, via dimensionality reduction, into something that captures the essence of the data. PCA allows us to {\it compress} the joint angle trajectories of the robot into a set of low dimensional trajectories, which are nothing but the kinematic motion primitives (kMPs).

The paper is organized as follows: Section \ref{sec:rl} introduces the Deep Reinforcement Learning (D-RL) framework for walking. Section \ref{sec:PCA} has a detailed  description on kinematic motion primitives (kMPs) and the associated kMP analysis of D-RL based gaits. Finally Section \ref{sec:results} describes the experimental results of Stoch walking.

%% file: RL.tex
\section{Reinforcement Learning Framework for Quadrupedal Walking}\label{sec:rl}

In this section, we will study the deep reinforcement learning (D-RL) framework used for realizing quadrupedal walking in simulation. We will also discuss the %associated challenges with the learning methods, and the
problems with transferability 
% issue from simulation 
to real hardware.

\subsection{Definitions}
We formulate the problem of locomotion as a Markov Decision Process (MDP). An MDP is represented by a tuple $\{S, A, P, R, \gamma \}$. Here $S \in \R^n$ is the set of robot states referred to as state space, and $A \in \mathbb{R}^m$ is the set of feasible actions referred to as the action space. $P: S\times A \times S\rightarrow [0,1]$ is the transition probability function that models the evolution of states based on actions, and $R: S \times A \rightarrow \R$ is the reinforcement or the reward obtained for the given state and action. $\gamma$ is called the discount factor defined in the range $[0,1]$. 

The goal in RL is to achieve a policy that maximizes the expected cumulative reward over time. A policy, denoted as $\pi: S \times A \to [0,1]$, is the probability distribution of the actions over the states. When dealing with huge state and action spaces, a parametrized policy is used, denoted as $\pi_\theta$, with the parameters $\theta$. %The optimal parameters and/or policy yields the maximum sum of the cumulative rewards. %Since the policy is a distribution, 
We use the following to obtain the optimal policy $\pi_{\theta^*}$: 
\begin{align}
\theta^* = \arg \max_\theta \sum\limits_{t=1}^{T} \E_{(s_t, a_t) \sim p_{\pi_\theta}(s_t, a_t)}[R(s_t, a_t)],
\end{align}
where $p_{\pi_\theta}$ is the marginalization of the state-action pair given the set of all possible trajectories $(s_0, a_0, s_1, a_1, ..., s_T)$ following policy $\pi_\theta$. The optimal parameters $\theta^*$ are evaluated iteratively by taking gradient ascent steps in the direction of increasing cumulative reward. More details on the algorithm, and the associated optimization technique are provided in Section \ref{algo}.

%The particular choice of action space can be easily bounded for the hardware safety, covers all permitted leg configurations and most importantly reduce the training time and improve the efficiency.
% \begin{figure}[h!]
% \includegraphics[width=\linewidth]{learning_arch1.jpg}
% \label{fig:training_loop}
% \caption{Reinforcement learning training framework}
% \end{figure}

\subsection{State and Action Space}

\label{sec:model}
%\section{Hardware Description}
We will briefly describe the quadrupedal model of Stoch.  
% All the four legs of the robot are identical and arranged in a symmetric fashion. 
There are four identical legs with two actuators per leg.
% and eight actuators in the robot which are distributed equally among its joints.
Each leg has a hip and a knee joint, and the actuators provide flexion-extension in each joint. Fig. \ref{fig:robot_pic} shows both the experimental and the simulated robot model.
% Each leg is identical, and has a total of 2 degrees of freedom: hip flexion-extension and knee flexion-extension. Hence, there are totally eight actuators in Stoch. 
Details on the experimental model are in Section \ref{sec:results} and \cite{dhaivatdesigndevelopment}.

\subsubsection{State Space}
The state is represented by angles, velocities, torques of the active joints, and body orientation (in quaternions). % is also included in the state space.
The combined representation yields a 28D state space. 
Note that the motor torques are typically treated as states in RL framework.
% This is a deviation from the traditional (control theoretic) definition of state space.
% (cite the google paper and the equation number where they included torque in the statespace).
% However, calculating velocities and accelerations is noisy on a physical system, therefore we are using joint-angles and base orientation (in quaternions) as the state space in our problem. 
\subsubsection{Action Space}
Reference \cite{action_space} demonstrated the effectiveness of choosing actuator joint angles as action space. The planar four bar linkage configuration of the legs, however, lead to self-colliding regions in the motor space. This creates a non-convex search space and degrades the learning efficiency. Hence, we chose to learn the legs' end-point position in polar coordinates represented as $\{r_i, \alpha_i\}$ where $i \in \{1,2,3,4\}$. The polar coordinates for the four legs collectively provide an $8$ dimensional action space. For each $r_i, \alpha_i$, we can obtain the joint angles (both hip and knee of leg $i$) via an inverse kinematics solver. We will denote the angles obtained from the inverse kinematics solver as $\{\beta_i^1, \beta_i^2\}$. % for each leg subjecting to the constraints to avoid internal collisions.   
% The learned end-point positions are fed into an inverse kinematic solver that provides the motor angles $\{\theta_i^1, \theta_i^2\}$ for each leg, to be executed by the robot. 
The polar coordinates can be restricted to a bounding box, thereby indirectly imposing angle limits on the joint angles.

\subsection{Network and Learning Algorithm} \label{algo}
Since the states and actions are continuous, we seek for algorithms that enable us to construct optimal policies that yield action values in the continuous space. Proximal Policy Optimization (PPO) \cite{PPO} have proved to be very successful for obtaining optimal continuous action values \cite{rlmatters}. It is an on-policy, model-free algorithm based on actor-critic learning framework. An actor (also called policy) decides an action given a state and a critic aims to improve the policy by evaluating the action value function. Both actor and critic networks are parameterized by neural networks with parameter $\theta$ and $\phi$ respectively. 
% 	Both of these neural networks consist of two fully connected layers with the first and second layers consisting of 200 and 100 nodes respectively. Activation units in the actor and critic networks are ReLU $\rightarrow$ ReLU $\rightarrow$ tanh, and ReLU $\rightarrow$ ReLU $\rightarrow$ linear respectively.

The learning algorithm samples $N$ episodes of maximum length $T = 1000$ using the current policy. The experience tuples consisting of $\{s_t, a_t, s_{t+1},r_t \}$, with $t \in [0, T]$, are stored and used for on-policy update of the parameters.
Policy gradient method \cite{advantage} gives the objective function to optimize the policy parameters given as:
% To optimize the policy parameters Sutton et al.\cite{advantage} provided the objective functions as:
\begin{align}
L^{PG}(\theta) = \E_{\pi_\theta} \Big[\log\pi_\theta(a_t|s_t)\hat{A_t}\Big]
\end{align}
where $\hat{A_t}$ is called the estimated Advantage function. It is given by $\hat{A_t} = r(s_t, a_t) + \gamma V_\phi(s_{t+1}) - V_\phi(s_t)$, where $V_\phi$ is the value function at any state. Furthermore, in order to account for the disparity between the behavioral policy $\pi_{old}$, and the current updating policy $\pi_\theta$, the advantage estimation is multiplied with the importance sampling ratio given by
\begin{align}
L(\theta) = \E_{\pi_{old}} \Bigg[\dfrac{\pi_\theta(a_t|s_t)}{\pi_{old}(a_t|s_t)}\hat{A_t}\Bigg].
\label{impotance_sampling}
\end{align}
% Lastly, in order to avoid huge and sudden variations in the policy during training TRPO \cite{TRPO} puts a constraint on the KL-divergence between the old and the new policy. However, the PPO algorithm simplifies the process by penalizing on the high KL-divergence and solving an unconstrained optimization. Hence the final objective for the PPO algorithm is given by
Lastly, in order to avoid huge and sudden policy variations during training, PPO penalizes on the high KL-divergence and solves an unconstrained optimization. Hence the final objective for the PPO algorithm is given by
\begin{align}
\begin{split}
L^{KL}(\theta) = \: &\E_{\pi_{old}} \Bigg[\dfrac{\pi_{\theta}(a_t|s_t)}{\pi_{old}(a_t|s_t)}\hat{A_t}\Bigg] - \\ & \beta_{KL} \E_{\pi_{old}}\big[ D_{KL} (\pi_{old}(a_t|s_t) || \pi_{\theta}(a_t|s_t)) \big]
\end{split}
\end{align}
where $\beta_{KL}$ is an adaptive penalty coefficient. Given a target KL-divergence, $\delta$, $\beta_{KL}$ for the next policy update is calculated as:
\begin{itemize}
\item if $D_{KL} < \delta/1.5, \beta_{KL} \leftarrow \beta_{KL}/2 $
\item if $D_{KL} > \delta/1.5, \beta_{KL} \leftarrow \beta_{KL} \times 2 $
\end{itemize}

We used the open source implementation of PPO by Tensorflow Agents \cite{tf_agents} that creates the symbolic representation of the computation graph. The implementation is highly parallelized and performs full-batch gradient ascent updates, using Adam \cite{adam} optimizer, on the batch of data collected from multiple environment instances.

\subsection{Simulation Framework}
We used Pybullet \cite{pybullet} simulator, built on top of Bullet3 physics engine, for a realistic model simulation. A three-dimensional computer-aided-design (CAD) model is developed using SolidWorks \cite{solidworks} to capture the kinematics and inertial properties of the robot. This model is transferred to Pybullet by using a Universal Robot Description Format \cite{roswebsite}
 (URDF) exporter. In addition, actual mass of all the links, actuator force limits, joint limits and kinematic-loop constraints of the flexural joints are measured and manually updated in the URDF file for a more realistic simulation. % (refer Fig. \ref{fig:robot_pic}). 

\subsection{Reward Function}
We designed a reward function that gives positive reinforcement with the increasing robot's base speed and simultaneously penalizes high energy consumptions. The agent receives a scalar reward after each action step according to the reward function
\begin{align}\label{eq:reward}
r_t = w_{vel} \cdot {\rm{sign}}(\Delta x_t) \cdot \max \{ | \Delta x_t |, 0.1 \} - w_E\cdot \Delta E .%|\tau_t\cdot \omega_t|\cdot \Delta t
\end{align}
Here $\Delta x_t$ is the difference between the current and the previous base position along the $x$-axis. %$\tau$ and $\omega$ are the motor torques and motor velocities respectively. $\Delta t$ is the simulation time step 
$\Delta E$ is the energy spent by the actuators for the current step,
and $w_{vel}$ and $w_E$ are weights corresponding to each term. $\Delta E$ is computed as
\begin{align}
\Delta E = \Sigma_{i=1}^4 ( | \tau^1_i (t) \cdot \omega^1_i (t) | + | \tau^2_i (t) \cdot \omega^2_i (t) | ) \cdot \Delta t,
\end{align}
where $\tau^1_i, \tau^2_i$ are the motor torques, and $\omega^1_i, \omega^2_i$ are the motor velocities of the $i^{th}$ leg respectively.

\subsection{Gait Generation and Simulation Results}
% \subsubsection{Gait Design}
The actor and critic network in the learning algorithm consists of two fully connected layers with the first and second layers consisting of 200 and 100 nodes respectively. Activation units in the actor and critic networks are ReLU $\rightarrow$ ReLU $\rightarrow$ $\tanh$, and ReLU $\rightarrow$ ReLU $\rightarrow$ linear respectively. Other hyper-parameters are mentioned in Table. \ref{tab:hyperparam}. The pybullet simulator is configured to spawn multiple agents for faster collection of samples---in our case 30 agents were used in parallel threads.

Quadrupedal animals typically have eight types of gaits \cite{owaki2017quadruped,Fukuoka2015}. Reference trajectories were used in \cite{DBLP:journals/corr/abs-1803-05580} open-loop signals were used in \cite{google_paper} to learn a specific gait. However, in our work, we learned everything from scratch. To learn a specific gait type, we leveraged on gait symmetries and placed hard constraints on the foot position. For e.g. in trot, diagonally opposite legs are in sync. Hence, we provide same $\{r, \alpha \}$ to  these legs. This also results in reduction in the dimensions of action space from $8$ to $4$. 
% To learn a specific gait type we put hard constraints on the legs' phase difference (e.g. in trot, diagonally opposite legs have the same phase resulting in reduction in the dimensions of action space from $8$ to $4$). 

The proposed approach yields walking gaits that are efficient, fast and also robust in simulation. The learning algorithm runs for a maximum of $15$ million steps and the observed training time is $4.35$ hours on a Intel Core i7 @3.5Ghz$\times 12$ cores and 32 GB RAM machine.

\begin{table}[t!]
\centering
\begin{tabular}{ c | c }
  Entity & Value  \\
  \hline
  $w_E, w_{vel}$ & $0.05, 1.0$  \\
  Discount factor $(\gamma)$ & $0.994$ \\
  Learning rate (actor, critic) & $10^{-4}, 10^{-4}$ \\
  Training steps per epoch & $3000$  \\
  KL target ($\delta$) & $0.01$  \\
  End-point position limit ($\theta$) & $[\ang{-30},\ang{30}]$\\
  End-point position limit ($r$) & $[15$cm, $24$cm$]$\\
\end{tabular}
\caption{\small Hyper-parameter values of the learning algorithm}
\label{tab:hyperparam}
\end{table}
% For the rest of the paper we will mainly use the trot gait as reference; but the approach has also been successfully used to generate other gaits.

%Using prior knowledge of the gaits, than restructuring the reward function to generate different gaits from scratch, speeds up the learning process. Therefore, we put hard constraints on the legs' phase difference (e.g. in pace, legs on each side have the same phase). This constraint effectively reduce the dimensions of the action space from $8$ to $4$.

\subsection{Challenges}

There are two main challenges in realizing D-RL based gaits in quadrupeds:

\subsubsection{Training for model and environment updates} The training framework has a large set of hyper-parameters (a small subset of them are shown in Table \ref{tab:hyperparam}) that require tuning in order to get desirable gaits. These gaits are, indeed, robust and efficient, but any update in the model or change in constraints in the gait requires retuning of these hyper-parameters in order to obtain desirable gaits. Retuning is not intuitive and requires a series of training iterations.
This is a common problem even in nonlinear model based walking optimizations \cite{reher2016algorithmic}, where small variations in constraints often lead to infeasible solutions.

\subsubsection{Transferability on real hardware}

Transferability of these learned gaits in the real hardware faces numerous challenges. Some of them are mainly due to uncertainties due to unaccounted compliances, imperfect sensors, and actuator saturations. In addition, inertial and electrical properties of  actuators vary from one joint to another. %These uncertainties often resulted in unrealizable gaits from the training. 
It can be observed that imposing very conservative constraints like tighter velocity and angle limits resulted in longer training times, which often converged to very unrealistic gaits. 

In order to put these challenges in context, it is necessary to review the methodology followed in \cite{google_paper}. They proposed to develop accurate robot model, identify the actuator parameters and study the latencies of the system. This procedure, indeed, resulted in more realistic gaits, but required multiple instances of training and is highly sensitive for any update in hardware. In addition, DNNs are computationally intensive (NVIDIA Jetson TX2 processor was used). Therefore, as a deviation from this approach, we primarily want to focus on a methodology that is more robust to hardware changes, and also computationally less intensive. As we add more and more tasks and objectives for a diverse set of environments, it is important to focus on using a {\it compressed} set of walking parameters with minimal training. 
% We propose a solution that uses kinematic motion primitives (kMPs) extracted from principal component analysis (PCA). 
% We will show that it is possible to reconstruct the walking gaits learned by the DNNs via a minimal set of trajectories, called the kinematic motion primitives (kMPs). 
We will discuss this in detail in the next section.

%% file: PCA.tex
\section{Realizing Behaviors through kMPs}\label{sec:PCA}

The kinematic motion primitives (kMPs) are invariant waveforms, sufficient to explain a wide variety of complex coordinated motions, including discrete (e.g., reaching for a target with one hand) and periodic behaviors like walking and running \cite{moro2012kinematic}. This concept of motion primitives arose from biological observations \cite{moro2012kinematic,moro2011human,moro2013horse}. In essence, it was observed that biological motions can be described to a good extent with the combination of a small number of lower dimensional behaviors \cite{sanger2000human, stephens2008dimensionality}.  %It was apparent from the analysis that even with a large number of inputs (joints) almost 90\% of the motion behaviors can be explained with a small group of patterns \cite{sanger2000human}, which were later known to be called as kMPs.
%In the early 2000s, 

The process of determining kMPs are primarily conducted by Principal Component Analysis (PCA) \cite{lim2005movement,moro2011human, moro2013horse,sprowitz2014kinematic,jolliffe2011principal}. %PCA is one of  the many dimensionality reduction techniques available in the literature. 
PCA is useful when there is extensive data from walking demonstrations, which can be  processed to extract the kMPs. These kMPs can then be adapted/transformed to realize walking in robots. %, and then adapted in the humanoid/quadrupedal robots \cite{moro2011human,moro2012kinematic,moro2013horse,sprowitz2014kinematic}. 
In essence, principal components yield vectors in the joint space with the highest variance for the given data.
% This results in minimal representation of the data with fewer vectors
A linear transformation of the joint space to these principal components results in dimensionality reduction. %and is simpler than other dimensionality reduction methods. 
A more detailed exploration of this analysis is done in \cite{lim2005movement} %via optimization, 
and \cite{kim2008fast}.% via motion generation. %, and \cite{kwon2008natural} via hidden markov models. %We will briefly describe the main methodology here, and then focus on extraction of kMP for the joint angle trajectories for our robot.

% \subsection{Main methodology.}
%authors first collect the data and study the  these behaviors to the humanoid robot\cite{moro2012kinematic} or quadruped robot\cite{moro2013horse}. The gait transfer between the horse/human and robots is conducted based on kinematic data alone. 

%In this work, we replace the source of data from a human/horse to the results of a Reinforcement learned(RL) policy, working in a simulator.   Implementation of RL policy in the simulator is advantageous. Since the learning is done on a realistic model of the robot, the transferring of the gait data does not require any human intervention. This makes it a formal method of learning transfer. With the extracted kinematic motion primitives and their weights, we control the robot with a CPG based controller on-board.

%In recent times, these studies resulted in development of control strategies for both humanoid robots. %In these works, the extracted lower dimension behavior form human/animal locomotion is refered to as  Kinematic Motion Primitive (kMP).  

We will first describe the methods used for the data analysis, and then the associated principal components.  % that are used to generate the kMPs.These kMPs are then used to reconstruct the trajectories to realize various gaits for Stoch. 
kMPs for the different gaits, trot, pace and bound, will be extracted and compared.
% Since we chose to realize trot in simulation, we will focus on extracting the kMPs from this gait. %Later on, we will show that the same kMPs can also be used to realize derived behaviors, namely walk and bound, which were not possible via D-RL based policies.
% Note that these kMPs can also be used to generate other kinds of gaits namely walk, bounding etc by simply varying the phase difference between the different joint angle trajectories.
% Towards the end of this section, We will provide a statistical analysis of all the gaits realized in Stoch. 
Towards the end of this section, we will also provide a statistical comparison between the different kMPs. %extracted from different gaits in simulation, and also the kMPs extracted from horse \cite{moro2013horse}.

% In section ??, we provide a statistical analysis of the xxxx gaits and a statistical comparison between the kMPs of a horse[19] and Stoch.

\subsection{Extraction of kMPs}

%In Section \ref{sec:rl}, we described the process of generating a control policy with the help of reinforcement learning. Once the learning process is complete, we test the control policy in a simulation environment, PyBullet. 
Having obtained walking in simulation based on a trained policy, we record the joint angles and body orientation of the robot for $4800$ steps. %The data contains a set of $8$ joint trajectories for each trial. % The number of valid data sets used in each experiment is reported in section \ref{sec:results}. 
%Now, we explain the methods of data extraction from the recorded data. 
The recorded data is then processed as follows: First, we remove front and rear $15$\% of the data set to avoid transient cases. Second, we divide the data into segments by peak detection. All these data segments are then shifted and scaled w.r.t. time in order to align the end points. By polynomial interpolation, we reconstruct trajectories in between the data points. With this reconstruction, we can directly obtain the mean data. % (average of the data at every time instant). 
%and transform it into a zero mean dataset. %Next, we conduct the Principal Component Analysis on this data. 

%\subsection{Extraction of motion primitives}

%The motion primitives are extracted via a principal component analysis of the joint angle data. 

\begin{figure}
\begin{minipage}[t]{.72\linewidth}
\vspace{5pt}
\centering
	\includegraphics[width=0.96\columnwidth]{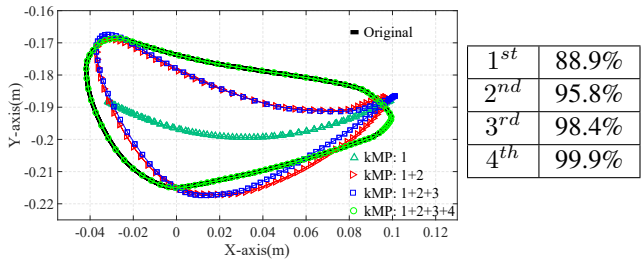}%
%       \caption{A figure}%
%   \label{fig:kMP_recon}
\end{minipage}%
\begin{minipage}[t]{.25\linewidth}
\vspace{7mm}
\centering
    \begin{tabular}{|c|c|}
	\hline
	$1^{st}$ &  88.9\% \\
	\hline
	$2^{nd}$ &  95.8\% \\
	\hline
	$3^{rd}$ &  98.4\% \\
	\hline
	$4^{th}$ &  99.9\% \\
	\hline
	\end{tabular}
%       \caption{Variance}%
%   \label{Table:kMP_PCV}
\end{minipage}
\caption{Figure on the left shows end point trajectories of the one of the feet after reconstruction from different numbers of kMPs. The X-axis corresponds to the forward direction, and the Y-axis corresponds to the vertical direction respectively. Table on the right shows the percentage cumulative variance for adding each component at a time. %The last row corresponds to four components being used for the final reconstruction.
}
\label{fig:kMP_recon}
\end{figure}

Given the mean data, let $N$ be the total number of points for each joint. Denote this data by $x[t] \in \mathbb{R}^{n_j}$, where $n_j$ is the number of actuated joints ($8$ for Stoch). We get a matrix of joint angle values as
% \begin{align}
$X = \left[ x[1] \quad x[2] \quad \dots  \quad  x[N] \right ]^T \in \mathbb{R}^{N\times n_j }.$
% \end{align}
%where $N$ is the total number of recorded data points. 
% We get the mean as
% \begin{eqnarray}
% \bar x & := & \frac{1}{N} \Sigma_{i=1}^N x[i] \nonumber \\
% \hat X & := & \left[ x[1]-\bar x \quad x[2] - \bar x \quad \dots  \quad  x[N]- \bar x\right ]^T, \label{eq:zero_mean}
% \end{eqnarray}
Let $\hat X$ be the zero mean data obtained from $X$. We compute the covariance as %= Recorded joint angle for the hip and knee,
\begin{eqnarray}
cov(\hat X)&=&\frac{1}{N}\hat{X}^T \hat{X}\label{eq:cov}
\end{eqnarray}
%Here $cov(\hat X)\in \R^{n_j \times n_j}$ is the covariance of $\hat X$ , $\bar{X}\in \R^{n_j}$ is the mean of $X$. 
% Let $\lambda_i$'s and $\hat{\textbf{e}}_i$'s be the eigenvalues and eigenvectors of $cov(\hat X)$. 
The principal components are precisely the eigenvectors, $\hat{\textbf{e}}_1, \hat{\textbf{e}}_2, \dots , \hat{\textbf{e}}_{n_j}$, of $cov(\hat X)$. Hence, for $8$ joint angles, we have $8$ principal components. Given the principal components, we have the $i^{th}$ kMP defined as % and $M_{P,i}$ is the .
%and the corresponding eigenvalues ($\lambda_i$) and eigenvectors ($\hat{\textbf{e}}_i$) of the  as
\begin{eqnarray}
%\left[cov(\hat X[1:N]) - \lambda_i  \textbf{I}\right] \hat{\textbf{e}}_i  &=& \textbf{0}\label{eq:Eig_vec}\\
M_{P,i} &:=& \|\hat{X}.\hat{\textbf{e}}_i\|_\infty^{-1} \hat{X}.\hat{\textbf{e}}_i, \label{eq:kMP_gen}
\end{eqnarray}
where $\|.\|_\infty$ (infinity norm) is used to normalize the magnitude. We typically use fewer components ($4$ for Stoch) for reconstructing the joint angle trajectories from the kMPs, since they account for over 99\% of the cumulative percentage variance (see table in Fig. \ref{fig:kMP_recon}).

Having obtained the kMPs, we reconstruct the original joint trajectories as follows. Denote the vector of kMPs as
% \begin{eqnarray}
$\mathbb{P} : = \begin{bmatrix} M_{P,1} & M_{P,2} & \dots & M_{P,n_c}  \end{bmatrix} \in \mathbb{R}^{N \times n_c}$,
% \end{eqnarray}
where $n_c$ is typically less than $n_j$ (for Stoch $n_c = 4$). We have the final reconstruction of the joint angles as
\begin{align}\label{eq:synergy}
\mathbf{Q} = \mathbb{P} \times \mathbf{S} + \mathbf{Z},
%\underbrace{\left[ \begin{matrix}
%q_1\\ \vdots \\q_{n_j}
%\end{matrix} \right]}_{\mathbb{Q}} &=& 
%\underbrace{\left[\begin{matrix}
%s_{11} & \cdots & s_{1j}\\
%\vdots & \ddots   & \vdots \\
%s_{i1} & \cdots & s_{ij}
%\end{matrix}\right]}_{\mathbf{S}} \times
% \underbrace{\left[\begin{matrix}
% M_{P,1}\\ \vdots \\M_{P,n_j}
% \end{matrix}\right]}_{\mathbb{P}}
% +\underbrace{\left[\begin{matrix}
% Z_1\\ \vdots\\ Z_j
% \end{matrix}\right]}_{\mathbb{Z}} 
%\nonumber\\
%or, \mathbb{Q}&=& \mathbf{S}\times \mathbb{P} + \mathbb{Z}
\end{align}
where $\mathbf{Q}= \mathbb{R}^{N \times n_j }$ is the matrix of desired joint angle values, %$\mathbb{P} \in \mathbb{R}^{N \times n_j}$ is the collection of kMPs generated from the data.
$\mathbf{S}  \in \mathbb{R}^{n_c \times n_j}$ is coefficient matrix, also called as kMP synergy map, that maps the kMPs to the joint space, and $\mathbf{Z} \in \mathbb{R}^{ N \times n_j}$ is the zero mean offset matrix (derived from $\bar x$), which is added back to the joint angles.

% \vspace{2pt}
 % added ending }

% \begin{figure}[h!]
% \centering
% \begin{tabular}{|c|c|}
% \hline
% $1^{st}$ &  88.9\% \\
% \hline
% $2^{nd}$ &  91\% \\
% \hline
% $3^{rd}$ &  96\% \\
% \hline
% $4^{th}$ &  99\% \\
% \hline
% \end{tabular}
% \caption{The percentage cumulative variance of one of the sample data.}
% \label{Table:kMP_PCV}
% \end{figure}

% \begin{figure}[h!]
% \centering
% \includegraphics[width=\linewidth]{kMP_end_effector.eps}
% \caption{The reconstruction of a joint trajectory with different number of Kinematic motion primitives(kMP) }
% \label{fig:kMP_recon}
% \end{figure}

% \subsection{Trajectory reconstruction and verification}
% This section focuses on the construction of joint trajectories starting from the kMPs. 

% \subsection{Kinematic Motion Primitives} \label{sec:results/kMP}

\subsection{Discussion \& Comparison}

\begin{figure}[t!]
\includegraphics[width=\linewidth]{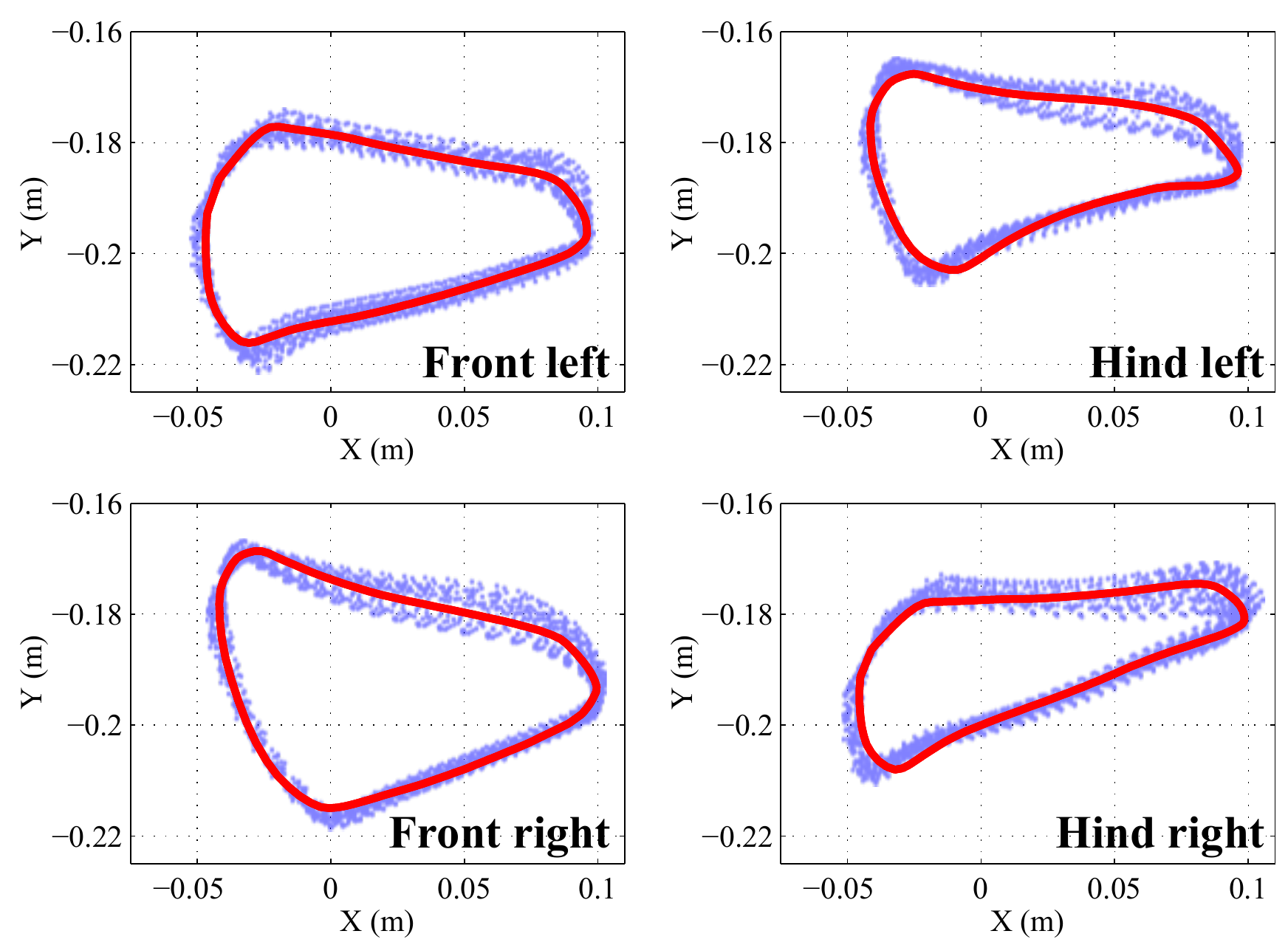}
  \caption{Figures showing the end point trajectories of all the four feet of the quadruped. The plots in blue correspond to the end point trajectories from simulation data. The plots in red are the reconstruction from the kMPs.}
  \label{fig:end point hardware}
\end{figure}

% We will first study the kMPs obtained from the simulation data, and then draw comparisons with the horse data. 
Fig. \ref{fig:kMP_recon} shows the result of reconstruction from the kMPs. A table showing the cumulative percent variance of the original trajectory with reconstruction from each kMP is also provided. It can be verified that only four kMPs were sufficient for the final reconstruction\footnote{This observation is true for any continuous trajectory that does not have ``high-frequency'' oscillations. This property was extensively used to reconstruct walking trajectories by using ellipses in \cite{kohl2004policy} and low-order basis functions in \cite{tedrake2004stochastic}.}.
Fig. \ref{fig:end point hardware} shows the comparison between the end foot trajectories obtained from the simulation, and from the reconstruction. It must be noted that kMPs yield trajectories that are as good as the time averaged trajectories of the original data (which is eight dimensional). In addition, with kMPs, there is existing literature for obtaining derived behaviors like walk, trot and gallop \cite{moro2012kinematic,sprowitz2014kinematic}, which will be explained more in the next section.
%, Table \ref{Table:kMP_PCV} \& figure \ref{fig:kMP_recon}. 

% In this section, we will discuss the aim of the kMP and the methods used to analyze \& compare the kMPs of different gaits, animals \& robots.

% To compare the extracted kMPs, a visual representation of the kMPs is provided for each experiment, together with related statistical information. The statistical comparison between two sets of kMPs is conducted via covariance and delay \cite{moro2012kinematic}. The statistical comparison  do not indicate any proof, it is a quantification of what we can already isualize\cite{moro2012kinematic}.

\begin{figure*}[t!]
\centering
\includegraphics[width=\linewidth]{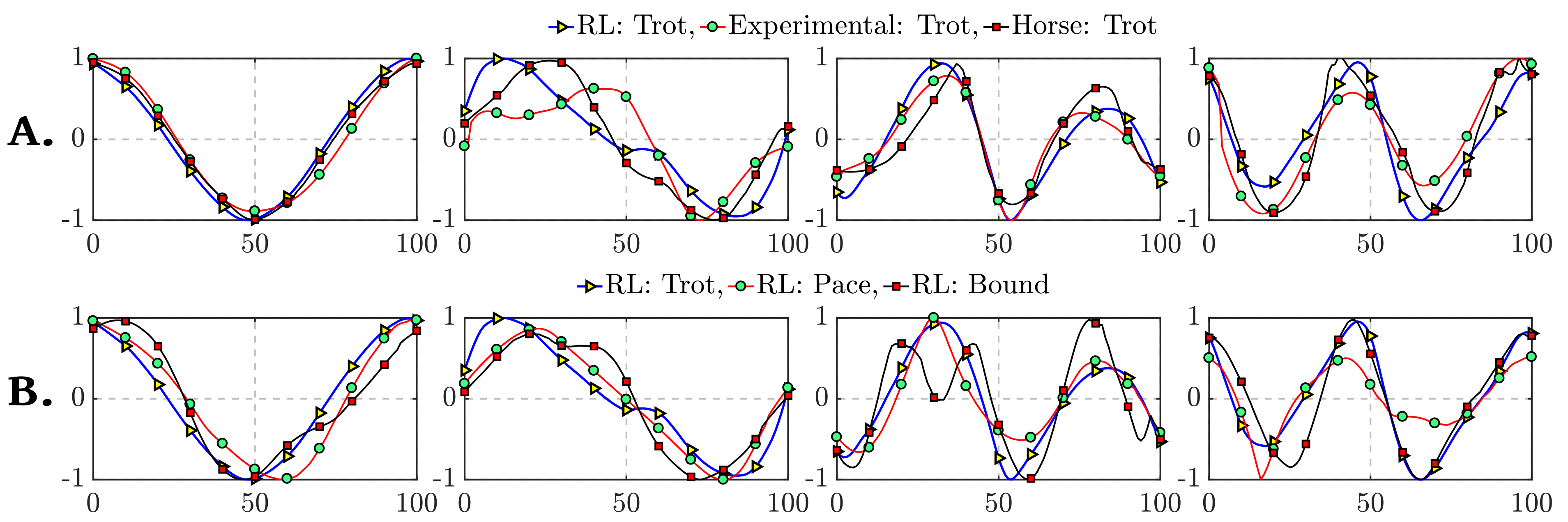}
    \caption{Here we make two types of comparisons. The top four figures (A) show the comparison between kMPs for the trot gait: hardware trot, simulation trot, and finally horse trot. The bottom four figures (B) show the comparison between kMPs obtained from the learned gaits in simulation: trot, pace and bound. The kMPs (of hardware, simulation or horse) seem to follow a pattern: \textbf{VNMW}. Statistical comparison of these patterns are provided in TABLE \ref{tab:cross_cov_HSH}}
  \label{fig:kMP_gait}
\end{figure*}

% Our analysis of the walking gait with the horse data is mainly motivated by the observations made by \cite{moro2013horse}. 
It was shown in \cite{moro2013horse} that biological quadrupeds (like horses) inherently contain some basic patterns that were consistent across different gaits and models. Based on this result, we compared the trot data obtained from our simulation with the trot data obtained from the horse, and the results are in Fig. \ref{fig:kMP_gait}. We observed that there was a high correlation not only between the kMPs obtained from the learned and horse gait (Fig. \ref{fig:kMP_gait}-A), but also between the kMPs obtained from different learned gaits (Fig. \ref{fig:kMP_gait}-B). Table \ref{tab:cross_cov_HSH} provides more statistical details, where it is shown that cross-covariance w.r.t. the first two kMPs is more than $90\%$. These observations point to the fact that even the kMPs derived from the RL based simulation followed very similar patterns, irrespective of the type of gaits and the model used.

%irrespective of the type of gait or the model used. 

% \begin{figure}[b!]
% \centering
% \includegraphics[width=\linewidth]{kMP_gait_compare.png}
% \caption{Figures showing the four kMPs for one step for the trot and pace gaits of the robot, which are in turn compared with the trot gait of the Horse kMPs. All of the kMPs show a very similar profile over the entire step. Horse kMP was extracted by digitizing the plots from \cite{moro2013horse}.
% % : Horse kMP \textbf{\color{red}(red, solid)}, Simulation kMP \textbf{\color{green}(green, dotted)}, Hardware kMP \textbf{\color{blue}(blue, dotted)}. The left most figure corresponds to the first component, and the right most figure to the fourth component. The figures are in the decreasing order of the eigenvalues.
% }
%   \label{fig:kMP_gait}
% \end{figure}

% More details are in the next section. %on experimental results based on the trajectories derived from $\mathbf{S}$ and $\mathbf{P}$'s.
%This requires almost no human intervention, and is, therefore, much more effective than other existing methods.

% It is important to note that this process requires refitting of the trajectories due to difference in the geometry and the mass properties of the humans/horse and the robots. 

% On the other hand, the data that we get from the simulator of Stoch requires no modification. 

% Therefore, our goal in this paper is to conduct a PCA of Stoch walking data obtained from the simulation, and then realize walking in Stoch experimentally via kMPs. 

%% file: Results.tex
\section{Experimental Results}\label{sec:results}

In this section we discuss the experimental results conducted on the quadrupedal robot Stoch. We first describe the hardware, and then describe the methodologies used to realize walking along with the results.

\subsection{Hardware Description of Stoch}
Stoch is a quadrupedal robot (see Fig. \ref{fig:robot_pic}) designed and developed in-house at IISc, Bengaluru \cite{dhaivatdesigndevelopment}. %As mentioned previously, Stoch has two joints and actuators for each leg. 
Model and actuator specifications are summarized in Table \ref{tab:specs_robot}. The legs are assembled with lightweight carbon fiber (CF) tubes for the linkages, and $3$D printed poly-lactic acid (PLA) parts for the joints. 
% The bearings at the joints are constructed by using custom made carbon fiber journals. %This feature reduces weight of the swing-leg. 
%The servo-motors (Kondo-$2350$HV, Robokits Ultra Torque: RKI-$1203$) are mounted in a PLA housing. %The main body is fabricated with carbon fiber tubing, 
%as strong and lightweight structural members, while the front and back plates (thick CF plates) form the rigid framing structure. 
% The body contains well protected all electronic parts, servo-motor as well coupled rotary encoders housing.
%For measuring the body orientation, we used a $9$ axis IMU (Sparkfun MPU $9150$). For measuring the joint angles, we used magnetic rotary (MR) encoders (Bourns EMS$22$a). %interfaced via SPI communication . 
PWM signals to the servo motors are sent via Adafruit $16$-Channel $12$-bit: PCA$968$.
%  attached to the custom designed power and distribution circuit board and interfaced via I2C communication. 
Main computation platform used is Raspberry Pi $3$-B. The electric power to the robot is supplied from Li-Po batteries. More details about the hardware are described in \cite{dhaivatdesigndevelopment}.
%($11.1$V(3s)$\times 2200$mAh for hip motors, and $7.4$V(2s)$\times 3000$mAh for knee motors).
% which are connected to the power distribution board via tether.  

 \begin{table}[h!]
 \centering
 \begin{tabular}{c c|c c}
 \hline
total leg length & 230 mm & min./max. hip joint & -45$^{\circ}$/ 45$^{\circ}$\\
leg segment length & 120 mm & min./max. knee joint & -70$^{\circ}$/ 70$^{\circ}$\\
total mass & 3.0 kg & max hip/knee speed & 461$^{\circ}$/s \\
max hip/knee torque & 32 kg-cm %& max knee speed & 461$^{\circ}$/s \\
%max knee torque & 32 kg-cm 
& motor power(servo) & 16 W \\

 \hline
 \end{tabular}
 \caption{Hardware specifications of Stoch. %Angle and velocity limits of the actuators (servo motors) are also shown.
 }
 \label{tab:specs_robot}
 \end{table}
 
% \textbf{Hardware Section Ends Here}

% \subsection{Trajectory reconstruction in hardware}

% Section \ref{sec:results/learning_gaits} describes the methods of generation of $trot$ and $pace$ gait with the use of Reinforcement Learning (RL).

% In Section \ref{sec:results/kMP}, we study the kMPs extracted from periodic motions of the simulation data. Section 3.4 shows an example of joint trajectory reconstruction from kMPs. In Section 3.5 an interpretation of the kMPs is given, showing the foot trajectory when it is partially reconstructed from single kMPs or a subset of the five kMPs. Finally, in Section 3.6, the application of the kMPs to generate a valid walking for the humanoid robot COMAN is presented.

% Section \ref{sec:results/HWexp} describes the results of the implementation of kMP in the hardware. 
\begin{table}[h!]
\centering
\resizebox{\columnwidth}{!}{%
\begin{tabular}{c|c||c|c|c|c||c|c|c|c}
\hline
\multicolumn{2} {c||}{} & \multicolumn{4} {c||} {\bfseries Cross-covariance} & \multicolumn{4} {c} {\bfseries Delay} \\
\hline
\multicolumn{2}{c||}{} & 1st & 2nd & 3rd & 4th & 1st & 2nd & 3rd & 4th\\
\hline
& RT-ET & 0.97 & 0.74 & 0.93 & 0.79 & 0 & -0.02 & 0.02 & 0.04\\
\textbf{A.} & RT-HT & 0.99 & 0.89 & 0.86 & 0.82  & 0 & -0.02 & -0.05 & -0.04 \\
& ET-HT & 0.99 & 0.77 & 0.88 & 0.87 & 0 & 0.15 & -0.08 & 0.03 \\
\hline
& RT-RP & 0.87 & 0.94 & 0.91 & 0.7 & -0.03 & -0.07 & -0.08 & 0.16 \\
\textbf{B.} & RT-RB & 0.81 &0.84 & 0.72 & 0.88 & 0 & -0.02 & -0.05 & -0.04 \\
& RP-RB & 0.96 & 0.92 & 0.85 & 0.72 & -0.01& 0.0 & -0.08 & 0.18 \\
\hline
\end{tabular}%
}
\caption{%\textbf{A.} shows the comparison between different kMPs. 
\textbf{RT,RB,RP} are trot, bound and pace gaits obtained from D-RL respectively. \textbf{ET} is the trot gait obtained from experiment, and \textbf{HT} is the trot gait obtained from the horse. %The delay between the two kMPs is also shown to be very small. 
\textbf{A.} (resp. \textbf{B.}) shows the comparison between \textbf{RT,ET,HT} (resp.   \textbf{RT,RB,RP}).
% : \textbf{RT} for trot, \textbf{RP} for pace, and \textbf{RB} for bound respectively.
}
\label{tab:cross_cov_HSH}
\end{table}

\subsection{Synergy Matrices}
The general procedure to obtain the walking gaits for Stoch is as follows. We first reconstruct the trot gait from the kMPs by using the synergy matrix $\mathbf{S}$. If $\mathbf{Q}_s$ is the trot data obtained from the simulation, then we obtain $\mathbf{S}$ as
\begin{align}
\mathbf{S} = \mathbf{P_{inv}} \times ( \mathbf{Q}_s - \mathbf{Z} ),
\end{align}
where $\mathbf{P_{inv}}$ is the pseudo-inverse of $\mathbf{P}$. 
% from \eqref{eq:synergy} by computing the left-inverse of $\mathbf{P}$. 
This is computed offline. Synergy matrices for other types of gait %, namely walk and bound, 
are obtained by modifying $\mathbf{Q}_s$. For example, to obtain the bound gait, we modify $\mathbf{Q}_s$ (by time shifting) in such a way that the front legs are in phase (resp. hind legs). We have also reshaped the end point trajectories in small amounts, and obtained the corresponding $\mathbf{Q}_s$.
Since we are manually modifying the data,
% Since $\mathbf{P_{inv}}$ is not unique, 
a better approach would have been to use optimization, %to synthesize these synergy matrices
which will be explored in future. %\footnote{A better approach would have been to use optimization to synthesize these synergy matrices, which will be explored in future work.}.

  %We can obtain the same behavior by obtaining the corresponding synergy matrix for bound. 
% Joint angle data for computing the synergy matrix is obtained by phase shifting of the trajectories. 

% Section \ref{subsubsec: data_compare} shows the comparison of the kMP between the trot gaits learned in robot simulation, implemented in hardware and extracted from horse. 

\subsection{Results}
Having obtained the kMPs, and the synergy matrices for trot, walk, gallop and bound gaits, we reconstruct the joint angle trajectories by using \eqref{eq:synergy} in hardware. These joint angle trajectories are then used as command positions to each of the actuators (see \cite[Section IV-B]{dhaivatdesigndevelopment}). Video results showing different types of walking, and robustness tests are shown in \url{https://youtu.be/kiLKSqI4KhE}
%\cite{robot_video}. 

Due to space constraints, we only show the experimental data for the trot and its derived gaits. We have also provided kMPs extracted directly from the experimental trajectory, and the results are in Fig. \ref{fig:kMP_gait}-A and Table \ref{tab:cross_cov_HSH}. It can be verified that experimental kMPs also have a high correlation with the horse and RL kMPs. % shows the statistical comparison for the same.  %This strongly indicates the ``naturalness'' of the gaits realized not only in simulation, but also in hardware. 
We have also provided Table \ref{tab:speed} to compare the speeds of the various gaits obtained.
%We have also realized gaits by varying the shape of the end point trajectories of trot, and then obtaining the corresponding synergy matrices. This is to show that kMPs are versatile to small modifications of the gait. More details are also in the video.

% The tables in this section report only the final results of the statistics applied. In section \ref{subsubsec: Gait_change}, we change gait from trot to pace and vice verse by just switching the elements of the weight matrix.
% In section \ref{subsubsec: Turning}, we generate turning motion by manipulating the weights of the weight matrix.
\begin{table}
\centering
\resizebox{1\columnwidth}{0.04\linewidth}{%
\begin{tabular}{|c|c|c|c|c|c|c|}
\hline
%  Gait type& RL Trot & RL Bound & E Trot & E Bound & E Gallop & E Walk & Modified Trot \\
  Gait type & Trot & Walk & Gallop & Bound & Modified Trot 1 & Modified Trot 2 \\
\hline
 Speed $(m/s)$ & 0.60 & 0.51& 0.51 & 0.55 & 0.62 & 0.59\\
 \hline
\end{tabular}%
}
\caption{Walking speed of Stoch obtained during  different gait experiments. The last two columns are the result of reshaping of the end point trajectories of trot. %All the gaits are shown in the video.
}
\label{tab:speed}
\end{table}

% \subsection{Robustness results}
% To study the robustness of Stoch walking, we introduced external disturbances like kicking, pushing. 

In order to put these observations in perspective, it is important to discuss the kMPs derived from horse data more. We cannot realize walking by using only horse kMPs. Based on \eqref{eq:synergy}, we will need to determine the right synergy matrix $\textbf{S}$ that suits the given set of kMPs. %This is an optimization problem, which will be explored in future. 
It is also important to note that the kMPs derived from horses may not be the optimal choice for our quadruped (there are infinitely many kMPs). On the other hand, due to the fact that we already have (locally) optimal trajectories obtained from training, 
% we can simply compute $\mathbf{S}$ from the simulation data.  %D-RL based gait generation, in essence, yields (locally) optimal gaits due to the reward function. 
% The simulation data is obtained from a much closer model, and, hence, 
we focus on computing the synergy matrix from simulation kMPs. With these kMPs we were able to demonstrate not only trotting, but also other derived behaviors like walking, galloping and bounding.

% \subsubsection{Gait generation}\label{subsubsec: Gait_change}

% \subsubsection{Turning}\label{subsubsec: Turning}